\def\year{2021}\relax
\documentclass[letterpaper]{article} 
\usepackage{aaai21}  
\usepackage{times}  
\usepackage{helvet} 
\usepackage{courier}  
\usepackage[hyphens]{url}  
\usepackage{graphicx} 
\usepackage{amssymb}
\usepackage{amsmath}
\usepackage{multicol}
\usepackage{multirow}
\usepackage{bbm}
\usepackage{subfigure}
\usepackage{natbib}  
\usepackage{caption} 
\newcommand{\eg}{\emph{e.g.}}
\newcommand{\ie}{\emph{i.e.}}

\title{Exploiting Learnable Joint Groups for Hand Pose Estimation}
\author {
        Moran Li\textsuperscript{\rm 1\thanks{Equal contributions}}, 
        Yuan Gao\textsuperscript{\rm 1,2\footnotemark[1]}, 
        Nong Sang\textsuperscript{\rm 1} \\
}
\affiliations {
    \textsuperscript{\rm 1} Key Laboratory of Image Processing and Intelligent Control, \\
    School of Artificial Intelligence and Automation, Huazhong University of Science and Technology, Wuhan, China \\
    \textsuperscript{\rm 2} Tencent AI Lab \\
    moran\_li@hust.edu.cn, ethan.y.gao@gmail.com, 
    nsang@hust.edu.cn
}
\begin{document}

\maketitle

\begin{abstract}
In this paper, we propose to estimate 3D hand pose by recovering the 3D coordinates of joints in a \emph{group-wise} manner, where less-related joints are automatically categorized into different groups and exhibit different features. This is different from the previous methods where all the joints are considered holistically and share the same feature. The benefits of our method are illustrated by the principle of multi-task learning (MTL), i.e., by separating less-related joints into different groups (as different tasks), our method learns different features for each of them, therefore efficiently avoids the negative transfer (among less related tasks/groups of joints). The key of our method is a novel binary selector that automatically selects related joints into the same group. We implement such a selector with binary values stochastically sampled from a \emph{Concrete} distribution, which is constructed using \emph{Gumbel softmax} on trainable parameters. This enables us to preserve the differentiable property of the whole network. We further exploit features from those less-related groups by carrying out an additional feature fusing scheme among them, to learn more discriminative features. This is realized by implementing multiple 1x1 convolutions on the concatenated features, where each joint group contains a unique 1x1 convolution for feature fusion. The detailed ablation analysis and the extensive experiments on several benchmark datasets demonstrate the promising performance of the proposed method over the state-of-the-art (SOTA) methods. 
Besides, our method achieves top-1 among all the methods that do not exploit the dense 3D shape labels on the most recently released FreiHAND competition at the submission date. The source code and models are available at \url{https://github.com/moranli-aca/LearnableGroups-Hand}.
\end{abstract}

\section{Introduction}
\noindent 3D hand pose estimation is essential to facilitate convenient human-machine interaction through touch-less sensors. Therefore, it receives increasingly interests in various areas including computer vision, human-computer interaction, virtual/augmented reality, and robotics. The input of 3D hand pose estimation differs with the touch-less sensors, ranging from a single 2D RGB image \cite{z&B2017,cai2018weakly,mueller2018ganerated,spurr2018cross,yang2019disentangling,CVPR2019_Mesh_inthewild}, stereo RGB images \cite{STB}, to depth maps \cite{moon2018v2v,Wan_2018CVPR_Dense3d,ge2018P2P,ge2018PointNet,du2019crossinfonet,xiong2019a2j}.  This paper considers 3D hand pose estimation from \emph{a single 2D RGB image}, as it can be easily acquired from an ordinary RGB camera, therefore being most widely applicable in practice. 

Nevertheless, estimating 3D hand pose from a single 2D image is ambiguous. This is because multiple 3D poses correspond to the same 2D projection as the depth information is eliminated. Fortunately, the valid hand poses lie in a space with much lower dimensions, which can be learned to alleviate the ambiguities in a data-driven manner by deep learning technologies. Specifically, recent deep learning methods exploit the relationship of the hand joints/keypoints holistically to recover the valid hand poses. Recent representative methods typically realize this idea by using a fully connected network (FCN) on all the 2D joints to recover the 3D poses \cite{z&B2017,spurr2018cross}

It is arguable that not all joints are (equally) related. For example, joints that are far away from each other are less related (\ie, less constrained by skeleton/nerves) than those that are closed. This suggests that learning shared features for all the joints and exploiting them by a FCN is less appropriate. More specifically, \emph{from multi-task learning (MTL) perspective, recovering the 3D coordinates of each joint corresponds to a single task, this suggests that using shared features for less related tasks may lead to negative transfer and therefore results in degraded performance}.

The above discussion motivates us to learn different features for less related joints. This can be implemented by separating the joints into different groups, where related joints are contained in the same group and share the same features, representing one task in MTL. However, being different from the standard MTL, the group where the related joints should be categorized into (\ie, the task) is unknown in 3D hand pose estimation. In this work, \emph{we learn the groups of related joints in an end-to-end manner by introducing a novel differentiable binary joint selectors, which is stochastically sampled from a Concrete distribution} \cite{maddison2016concrete}. 

\begin{figure*}[t]
\begin{center}
\includegraphics[width=0.9\linewidth]{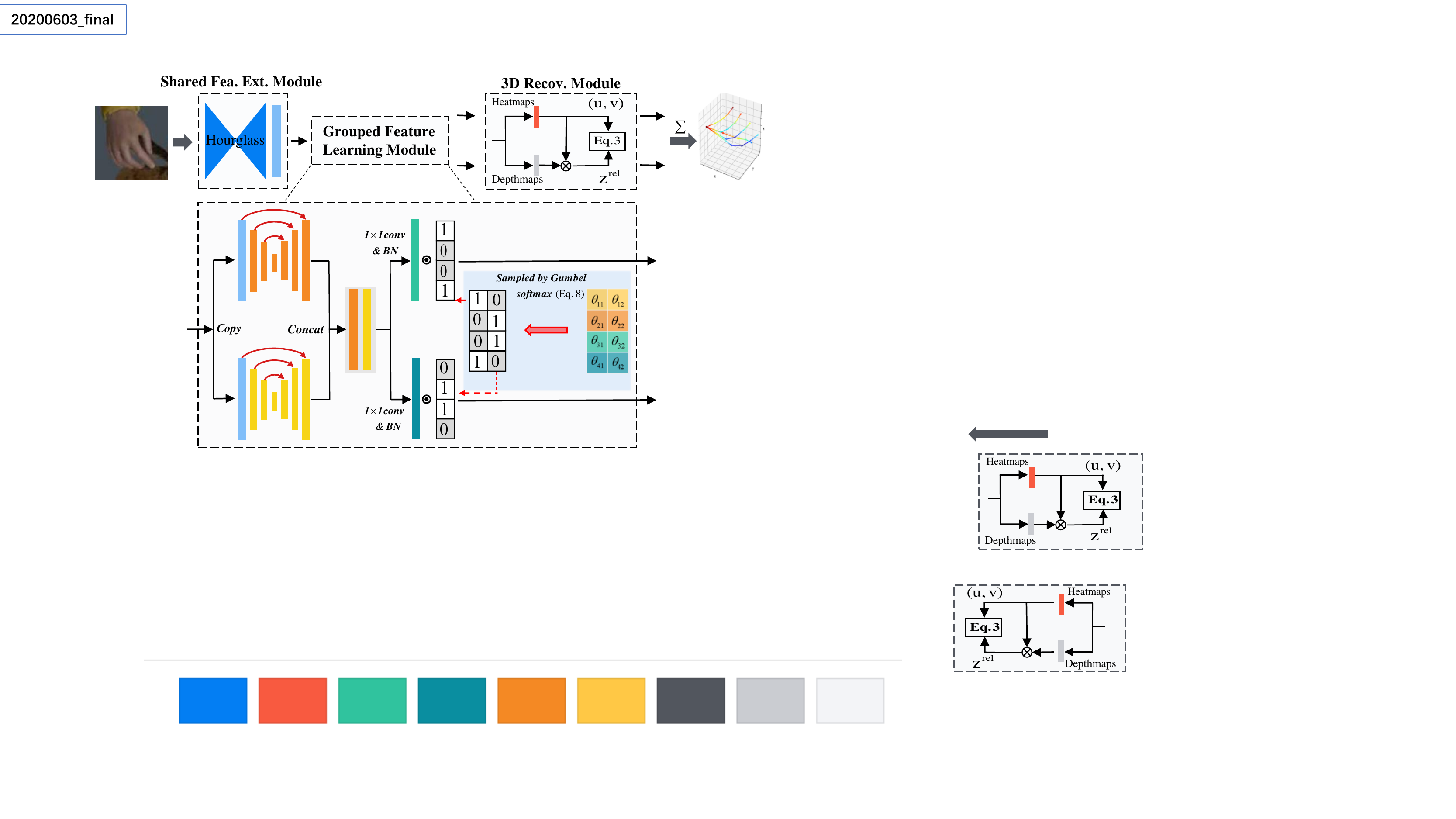} 
\end{center}
\vspace{-2mm}
  \caption{Overview of the proposed method. Our model consists of the \emph{Shared Feature Extraction Module, Grouped Feature Learning Module}, and \emph{3D Recovery Module}. $\odot$ denotes the element-wise product, and $\theta_{i,j}$'s are learnable network parameters for the binary selectors. We illustrate two groups for conciseness, while our method is able to categorize the joints into an arbitrary number of groups (we use $\#\text{Group}= 3$ in most of our experiments). (Best view in colors.)}
\label{fig:overall}
\end{figure*}

Moreover, features from different groups can be further exploited in our method without introducing undesirable negative transfer. To do that, \emph{we implement a feature embedding motivated by \cite{gao2019nddr}, which learns what to share automatically to exploit the useful features in an embedded feature space}. The proposed feature embedding scheme is formulated by feature concatenation and 1-by-1 convolution, which can also be trained end-to-end. The full design of our method is illustrated in Fig.\ref{fig:overall}.
In summary, the contributions of the proposed method include:
\begin{itemize}
    \item We consider 3D hand pose estimation from a single RGB image as a multi-task learning problem, where we group related joints as one task and learns different features for different groups/tasks to avoid the negative transfer.
    \item We learn the groups of joints (i.e., the tasks) automatically by end-to-end trainable binary joint selectors. We learn such binary joint selectors by stochastically sampling from a \emph{Concrete} distribution \cite{maddison2016concrete}, which is constructed by performing \emph{Gumbel softmax} \cite{jang2016categorical} reparameterization on trainable continuous parameters.
    \item We further exploit features from different groups/tasks without introducing undesirable negative transfer, by learning feature embeddings among different groups also in an end-to-end manner.
\end{itemize}

The remaining of this paper is organized as follows. We first discuss the related works. Then, the details of the proposed method are presented, followed by the implementation details, experiment results, including detailed ablation analysis. Finally, we give conclusions. The FreiHAND competition including root recovery and qualitative results for the benchmark datasets are included in the supplementary materials.


\section{Related Works \label{related_works}}
\noindent{\textbf{3D Hand Pose Estimation from a Single RGB Image.}}
Current works in this area can be roughly categorized into direct regression methods including single-stage methods \cite{spurr2018cross,yang2019disentangling,AligningHandICCV2019} that directly regress 3D joints locations from single RGB images, two-stage ways \cite{z&B2017,cai2018weakly,mueller2018ganerated,Doosti_2020_CVPR} that first regress 2D joints locations then lift 2D to 3D , and latent 2.5D estimation \cite{iqbalLatent25D,spurr2020weakly}. Besides, some recent works focused on shape estimation \cite{malik2018deephps,CVPR2019_Mesh_inthewild,ge20193dMeshGCN,baek2019pushingShape,hasson2019learning,zhang2019ICCV_endtoendMesh,FreiHAND2019,dkulon2020cvpr,Baek_2020_CVPR}. These methods generally combine discriminate methods with some generative methods (\eg, the MANO \cite{mano_2017}) to improve generalization. Although the rich prior information (\eg, the geometrical/biological dynamic constraints of hands joints) embedded in the generative model can assist discriminate model learning, such methods usually need some extra expensive shape annotations and iterative refinements during training which are not efficient enough and needed careful initialization. All of those methods treat all joints equally and view joints estimation as a single-task problem. In this work, we treat joints regression from the multi-task perspective for the intuition that different joints play different roles and posse different interrelationships.  

\noindent{\textbf{Multi-task Learning.}}
Many recent works in detection, semantic segmentation, and human pose estimation have adopted MTL methods to boost performance with auxiliary labels or tasks \cite{Liang_2019CVPR_Detection_MTL,Lee_2019CVPR_Detection_MTL,Pham_2019CVPR_SemanticInstance_MTL}. Misra et al. \cite{misra2016cross} implemented a cross-stitch unit for cross tasks features sharing which requires elaborate network architecture design. GradNorm \cite{chen2017gradnorm} automatically balanced the loss of different tasks during training to make the network focus more on difficult tasks. Liu et al. \cite{Liu_2019CVPR_MTL} used a soft-attention module for shared feature selection with dynamic tasks-specific weights adjustments needing extra parameters. Gao et al. \cite{gao2020mtl,gao2019nddr} proposed an end-to-end “plug-and-play” $1\times1$ convolution embedding layer to better leverage all different tasks features with a negligible parameter increase. In this work, we treat the hand joints estimation problem from the MTL view. Since the difference among different hand joints is not as large as that among different tasks of the general MTL problems, the dynamic loss adjustment methods or some complex network architecture design are not appropriate enough for our problem. We adopt the lightweight and effective MTL methods proposed by \cite{gao2019nddr}.

\noindent{\textbf{Joints Grouping.}}
Since, arguably, not all joints are equally related, some recent works \cite{madadi2017end,chen2019poseREN,du2019crossinfonet,zhou2018hbe,YingWuCVPR2019} have implemented manually designed grouping. Du et al. \cite{du2019crossinfonet} divided hand joints into the palm and fingers groups considering much more flexibility of those fingers. Zhou et al. \cite{zhou2018hbe} grouped hand joints into the thumb, index, and other fingers for the reason that the combination of thumb and index finger can generate some gestures without other fingers. These manually designed group strategies have some limitations: one is that different people have different intuitions for joins grouping with different explanations; the other is that it is hard to decide which grouping way is better without extra comprehensive experiments. More importantly, the manual grouping way only considers the bio-constrains/structure of hands and does not include the dataset-dependent information. Tang and Wu \cite{YingWuCVPR2019} proposed a data-driven way to calculate joints spatial mutual information and using spectral clusters to generate joints groups, but this method needs extra pre-statistic analysis for each dataset and does not include the bio-structure of hands. Hence, we propose a novel automatically learnable grouping method to implicitly learn the dataset-dependent and bio-structure dependent grouping. Our method can avoid human intuition ambiguity and extra pre-analysis.

\section{Method \label{method}}
\noindent In this paper, we treat the 3D coordinates regression of multiple hand joints as a multi-task learning problem. By categorizing the most related joints into a group, we can formulate multiple groups as multiple tasks. This enables us to learn a unique feature set for each group, which avoids the potential negative transfer when exploiting a shared feature for all the joints \cite{z&B2017,spurr2018cross,mueller2018ganerated,cai2018weakly,iqbalLatent25D,yang2019disentangling,CVPR2019_Mesh_inthewild,ge20193dMeshGCN,zhang2019ICCV_endtoendMesh}.

In the following, we first give an overview of the proposed method. Then, we detail the design of our grouped feature learning module, which automatically groups most related joints without violating the end-to-end training. The proposed grouped feature learning module further enables learning a discriminative feature embedding to exploit features from different groups. Finally, we give the loss functions to train the entire network.

\subsection{Overview}
Our method consists of three modules, \ie, a shared feature extraction module; a grouped feature learning module that categorizes the joints into multiple groups, and learns a unique feature set for each of them, to avoid negative feature transfer across groups; a 3D joints recovery module that regresses the 2D location and the relative depth of each joint, and finally recovers the 3D joint coordinates using the intrinsic camera parameters. We choose a similar design of the shared feature extraction module and the 3D joints recovery module to those used in \cite{iqbalLatent25D}, which benefits from i) leveraging well-developed 2D joints estimation \cite{Hourglass2016}, and ii) estimating relative depth between joints instead of much more difficult absolute depth. Note that the core of our method, \ie, the grouped feature learning module, is general and can be applied to most (if not all) hand pose estimation models, such as \cite{ge20193dMeshGCN,CVPR2019_Mesh_inthewild,zhang2019ICCV_endtoendMesh,AligningHandICCV2019}.

The architecture of our method is shown in Fig. \ref{fig:overall}. We detail the design of the shared feature extraction module and the 3D joints recovery module in this section. We leave the core of our method, \ie, the general-applicable grouped feature learning module, discussed in the next section.

\noindent{\textbf{Shared Feature Extraction Module.}} The hourglass network \cite{Hourglass2016} illustrates the great potential in extracting features that are especially suitable to represent the joints/keypoints \cite{IntegralLoss,Wan_2018CVPR_Dense3d,ge20193dMeshGCN,zhang2019ICCV_endtoendMesh}. It is built on symmetric encoders and decoders. We use the hourglass network to extract the shared features from the input image, which are then fed to the grouped feature learning module to learn the unique features for each group of joints.

\noindent{\textbf{3D Joints Recovery Module.}} The group-unique features learned from the grouped feature learning module are fed to the 3D joints recovery module to recover the 3D coordinates of each joint. 

Specifically, the group-unique features are first decoded to learn the 2D heatmaps and the (relative) depth maps simultaneously. Then, the 2D coordinates and the relative depth of \textit{each joint} are obtained at the maxima location of each heatmap channel.
After that, the 2D coordinates and the (relative) depth of each joint are exploited to recover the 3D joints in the camera coordinate, by using the intrinsic camera parameters and a depth root.  
Specifically, our goal is to estimate the 3D joint coordinates $(x_i, y_i, z_i)$ from the 2D estimated coordinates $(u_i, v_i)$ and the estimated relative depth $z_i$. Following the configuration of \cite{z&B2017,cai2018weakly,spurr2018cross,iqbalLatent25D,yang2019disentangling,AligningHandICCV2019}, we assume the global hand scale $s_0$, the depth of the root joint $z_{root}$, and the perspective camera intrinsic parameters $    M = \begin{bmatrix}
     f_x & 0 &  p_x \\
     0 & f_y & p_y \\\
     0 & 0 & 1 
     \end{bmatrix}$ are known\footnote{Typically, the camera intrinsic parameters $M$ can be obtained by the EXIF information of the image, and we can (almost) always recover the global hand scale $s_0$ and the depth of the root joints $z_{root}$ by the Procrustes alignment between the estimations and the ground-truth \cite{schonemann1966generalized}.}, where $f_x, f_y$ are the focal lengths, $p_x, p_y$ are the principal point coordinates of the camera. Therefore, we have the following relationship:
\begin{equation}
[x_i, y_i, z_i]^T = s M^{-1} [u_i, v_i, 1]^T,
\label{P2J}
\end{equation}
where $s$ is a scalar to balance the equation, and we have $s = z_i$ from the third equation of Eq. \eqref{P2J}.

In addition, $z_i$ and $z_i^{rel}$ are only up to a global translation $z_{root}$ and a global hand scale $s_0$:
\begin{equation}
    z_i = s_0  z_i^{rel} + z_{root}, \label{relationship}
\end{equation}
Therefore, we have:
\begin{equation}
[x_i, y_i, z_i]^T = (s_0  z_i^{rel} + z_{root}) M^{-1} [u_i, v_i, 1]^T.
\label{final}
\end{equation}

In the next section, we will detail how to learn a unique feature set for each joint group which finally decodes to the 3D coordinates.
\subsection{Grouped Feature Learning Module}
We treat the 3D hand pose estimation as a multi-task learning problem by categorizing the joints into different groups as multiple tasks, where we construct multiple network branches to learn a unique feature set for each group, therefore efficiently avoid the negative transfer across different tasks (\ie, joints groups). 

\begin{figure}[ht]
\centering
\includegraphics[width=0.45\textwidth]{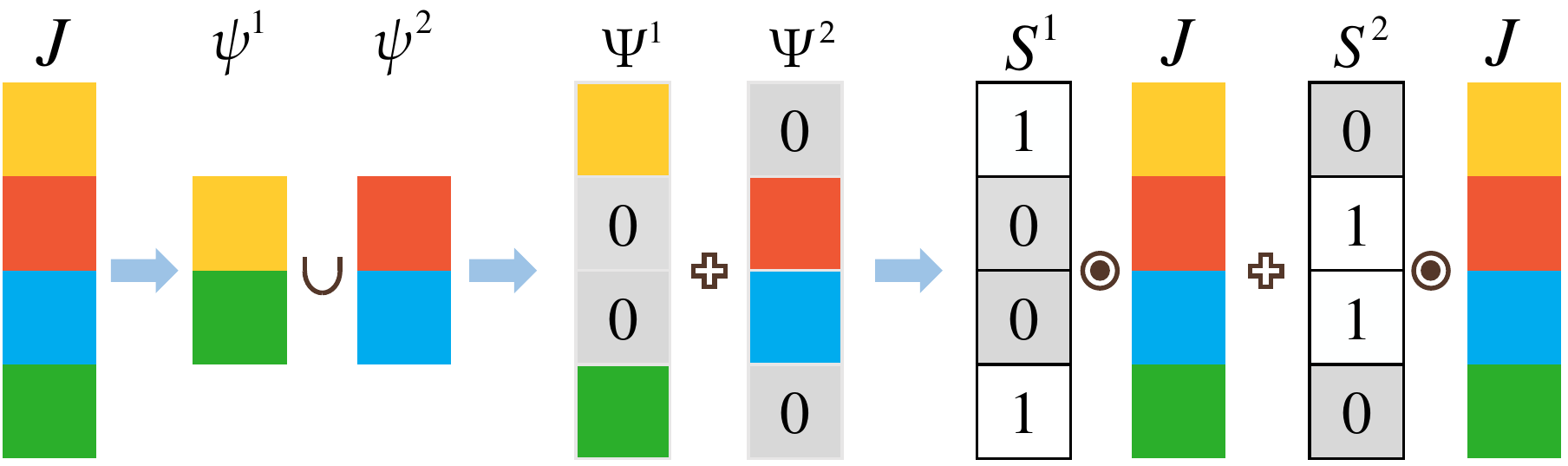}
\vspace{-2mm}
\caption{A simple illustration of our grouping operation. We show our grouping operation on four joints with two groups for conciseness. $\odot$ denotes the element-wise multiplication. The gray and the white blocks denote 0 and 1, respectively. (Best view in colors.)}
\label{fig:Grouping}
\end{figure}
To avoid ambiguities of group construction by different researchers \cite{zhou2018hbe,du2019crossinfonet}, we learn the best group formation in a fully data-driven manner. This is implemented by learning several binary selectors so that each selects one group of joints. We restrict that each joint should be selected exactly once, \ie, every joint should be selected to construct the groups, and different groups do not contain overlapped joints.
Formally, we aim to group $N$ joints into $K$ groups, denoted as:
\begin{equation}
    J = \bigcup_{k=1}^K \psi^k \quad \text{and} \quad \forall{(i ,j)}, \psi^i \cap \psi^j = \emptyset, \label{group1}
\end{equation}

\noindent where $J = \{(x_i, y_i, z_i)\}^N_{i=1} \in \mathbb{R}^{N \times 3}$ is the coordinates of $N$ joints, $\psi^k \in \mathbb{R}^{G_k \times 3}$ is the coordinates of $G_k$ joints in the $k^{th}$ group with $\sum_{k=1}^K G_k = N$, $\emptyset$ is the null set.
For ease of presentation, we augment each $\psi^g$ with 0's, as shown in Fig. \ref{fig:Grouping}, so that the augmented matrix $\Psi^g \in \mathbb{R}^{N \times 3}$ has the same dimension as $J$. Therefore, Eq. \eqref{group1} can be rewritten as:
\begin{equation}
    J = \bigcup_{k=1}^K \psi^k = \sum_{k=1}^K \Psi^k = \sum_{k=1}^K S^k \odot J, \label{group2}
\end{equation}
where $\odot$ denotes element-wise product, $S^k = \{ S^k_1, ..., S^k_N \} \in \mathbb{R}^{N \times 1}$ is a column vector representing the binary joints selector. Denoting $\mathbbm{1} \in \mathbb{R}^{N \times 1}$ as the column vector with all 1's, the element of $S^k$, \ie, $S^k_i$, has the following property:
\begin{equation}
\begin{aligned}
    S^k_i = \begin{cases}
    1  & \text{if } J_i \in \Psi^k \\
    0  & \text{if } J_i \notin \Psi^k
    \end{cases} \quad \text{and} \quad \sum_{k=1}^K S^k = \mathbbm{1}.
\end{aligned}
\label{constraints}
\end{equation}

We learn each $S^k$ under constraints of Eq. \eqref{constraints} as trainable model parameters of a neural network. To do that, we further construct a matrix $\mathcal{S} \in \mathbb{R}^{N \times K}$ and reformulate the constraints accordingly as:
\begin{equation}
\begin{aligned}
    &\mathcal{S} = [S^1, ..., S^K] \in \mathbb{R}^{N \times K}, \\
     \text{ s.t. } & \quad \forall i, \sum \mathcal{S}_{i, \cdot} = 1,  \text{and }  \mathcal{S}_{i,j} \in \{0, 1\},
    \label{constraints_final}
\end{aligned}
\end{equation}
where $\mathcal{S}_{i,j}$ is the element in $i$-th row and $j$-th column of $\mathcal{S}$, and $\mathcal{S}_{i, \cdot}$ is the $i$-th row of $\mathcal{S}$.

Equation \eqref{constraints_final} shows that each row $\mathcal{S}_{i, \cdot}$ follows a \emph{categorical} distribution. While we can easily impose the first constraint of Eq. \eqref{constraints_final} with a \emph{softmax} operation, the second constraint which binarizes each element $S_{i,j}$ violates the differentiable property of the whole network. 

To alleviate this issue, we reparameterize the \emph{categorical} distribution of $\mathcal{S}_{i, \cdot}$ using a \emph{Concrete} distribution to get the \emph{continuous relaxation} of $\mathcal{S}_{i, j}$ \cite{maddison2016concrete}, denoted as $\tilde{\mathcal{S}}_{i, j}$:
\begin{equation}
    \tilde{\mathcal{S}}_{i, j} \sim \frac{\text{exp}( \theta_{i, j}  + O_{i, j}) / \tau}{\sum_{k=1}^{K} \text{exp}( \theta_{i, k}  + O_{i, k} ) /\tau},
    \label{concrete}
\end{equation}
where $\theta_{i, j}$ is a learnable parameter of the network, representing the \emph{logits} of $\tilde{\mathcal{S}}_{i, j}$. $O_{i,j} = -\log (-\log (U_{i, j}))$ is the \emph{Gumble} variable with $U_{i, j} \sim (0, 1)$ as a \emph{uniform} distribution. $\tau = \tau(\text{training step})$ is the temperature parameter which is annealed to 0 with the training proceeds. It is shown in \cite{maddison2016concrete} that the continuous relaxation in Eq. \eqref{concrete}, i.e., the \emph{concrete} distribution, smoothly approaches to the \emph{categorical} distribution when the temperature $\tau$ approaches to 0. Therefore, we can sample $ \tilde{\mathcal{S}}_{i, j}$ from Eq. \eqref{concrete} as a good approximation for the binary selector $\mathcal{S}_{i,j}$.

The above analysis enables us to automatically categorize $N$ joints into $K$ groups, so that we can construct $K$ network branches, where each branch learns a unique feature set for each group. This efficiently alleviates the negative transfer across groups.

\noindent \textbf{Feature Fusing Across Groups.} We show that the different features learned by different groups can be further exploited to obtain more discriminative features, without introducing negative feature transferring. This is inspired by \cite{gao2019nddr} via learning multiple discriminative feature embeddings on the concatenated features from all the groups.

Specifically, denoting features from the $l$-th layer of the $k$-th group as $\mathbf{F}_l^k \in \mathbb{R}^{B\times H \times W \times C}$, where $B, H, W, C$ are the batch size, the height, the width, and the number of channels of the feature map, we can learn a more discriminative feature embedding $\hat{\mathbf{F}}_l^k \in \mathbb{R}^{B \times H \times W \times C}$ for group $k$ exploiting features from all the groups:

\begin{equation}
\begin{aligned}
\hat{\mathbf{F}}_l^k & = \texttt{BatchNorm} \Big( [\mathbf{F}_l^1, ..., \mathbf{F}_l^K] \cdot [\alpha_1^k \mathbf{I}_C, ..., \alpha_K^k \mathbf{I}_C]^{\top} \Big)\\
 & \xleftarrow{\text{Init. }\texttt{1x1conv}} \texttt{BatchNorm} \Big( \texttt{1x1conv} \big( [\mathbf{F}_l^1, ..., \mathbf{F}_l^K] \big) \Big),
\end{aligned}    
\label{nddr}
\end{equation}

\noindent where $[\mathbf{F}_l^1, ..., \mathbf{F}_l^k] \in \mathbb{R}^{B\times H \times W \times KC}$ is the concatenated feature from all the groups at layer $l$, and $[\alpha_1^k \mathbf{I}_C, ..., \alpha_K^k \mathbf{I}_C]^{\top}\in \mathbf{R}^{KC \times C}$ (with $\mathbf{I}_C \in \mathbf{R}^{C \times C}$ as the identity matrix) \emph{is a (reshaped) learnable 1x1 convolution with size $KC \times 1 \times 1 \times C$, which generalizes the weighted sum of $K$ features}.

In order to learn a good $\hat{\mathbf{F}}_l^k$ without introducing negative transfer, initially, it is reasonable to rely more on the original feature $F^k$ for the same group $k$, and the features from other groups $F^{i \neq k}$ also count for feature fusion, but with much smaller initial weights. Therefore, we carefully initialize $\alpha_k^k = 0.9$, which is the ``generalized weighted-sum'' weight for the original $\mathbf{F}_l^k$. We initialize the remaining $\alpha_{i \neq k}^k = (1-\alpha_{k}^k) / (K-1)$ for $F^{i \neq k }_l$, so that ensuring $\sum_{i=1}^K \alpha_i^k = 1$. This was also adopted in \cite{gao2019nddr}.
\subsection{Loss}
We use $\ell_1$ losses to train the network \cite{IntegralLoss,iqbalLatent25D,spurr2020weakly}. Denoting our estimation and the ground-truth 3D coordinates of the joints in the camera coordinate as $[X, Y, Z]^T = \{ (x_i, y_i, z_i) \}^{N}_{i=1}  \in \mathbb{R}^{N \times 3}$ and $[\hat{X}, \hat{Y}, \hat{Z}]^T = \{ (\hat{x_i}, \hat{y_i}, \hat{z_i}) \}^{N}_{i=1} \in \mathbb{R}^{N \times 3}$, respectively. We introduce a hyperparameter $\beta$ to balance the gradient magnitudes of XY and Z loss, hence the training loss is:
\begin{equation}
    L = || [X, Y] ^\top - [\hat{X}, \hat{Y}]^\top ||_1 + \beta ||Z - \hat{Z}  ||_1 \label{loss1}
\end{equation}

\section{Experiment \label{exp}}
\subsection{Datasets and Protocols}
{\bf RHD} (Rendered Hand Pose Dataset) \cite{z&B2017} is a synthesized rendering hand dataset containing 41,285 training and 2,728 testing samples. Each sample provides an RGB image with resolution $320\times 320$, hand mask, depth map, 2D and 3D joints annotation, and camera parameters. This dataset is built upon 20 different characters performing 39 actions with large variations in hand pose and viewpoints.

\noindent {\bf STB} (Stereo Hand Pose Benchmark) \cite{STB} is a real hand dataset containing 18,000 stereo pairs samples. It also provides RGB images with resolution $640\times480$, depth images, 2D and 3D joints annotations, and camera parameters for each sample. This dataset contains 6 different backgrounds, and each background has counting and random pose sequence. For each sample, we use one of the pair's images since the other contains the same pose and in almost the same viewpoint.  We split this dataset into a training set with 15,000 images and an evaluation set with 3,000 images following \cite{z&B2017}.

\noindent {\bf Dexter + Object} (Dexter and Object) \cite{sridhar2016real} is a real hand object interaction dataset consisting of six sequences with two actors (one female). For each sample, it provides an RGB image with resolution $640 \times 480$, depth image, camera parameters, and 3D annotation only for fingertips and three cubic corner joints of each object. We use this dataset as a cross-dataset evaluation similar to \cite{z&B2017,zhang2019ICCV_endtoendMesh}.

\noindent {\bf FreiHAND}  \cite{FreiHAND2019} is the latest released real hand dataset containing 130,240 training and 3,960 testing samples. Each training sample includes an RGB image with resolution $224 \times 224$, a 3D dense mesh ground-truth, and 3D joints annotations with hand scale and the camera parameters. The annotations of testing samples are not provided and the evaluation is conducted by submitting predictions to the online evaluation system. A lot of pose samples with severe occlusions contained in the FreiHAND makes it more challenging than other benchmark datasets.

\noindent {\bf Evaluation Protocols} We use the common metrics to evaluate the accuracy of the estimated 3D hand poses including mean/median end-point-error (3D mean/median EPE), the area under the curve (AUC) of the percentage of correct keypoints (PCK) with different thresholds. We assume that the global hand scale and the root joint location are known for the RHD and the STB datasets, following a similar condition as used in \cite{z&B2017,cai2018weakly,spurr2018cross,iqbalLatent25D,yang2019disentangling,AligningHandICCV2019}. For experiments on the Dexter + Object dataset, we follow the same way as Yang et al. \cite{yang2019disentangling}. On the FreiHAND dataset, only the hand scale and camera parameters are given for the testing samples. To be consistent with previous works, we adopt the root recovery method similar to Spurr et al. \cite{spurr2020weakly} to generate absolute root joints depth (see Supp.).

\begin{figure*}[htb]
\centering  
\subfigure[RHD]{
\label{Fig.AUC_Sota.1}
\includegraphics[width=0.3\textwidth]{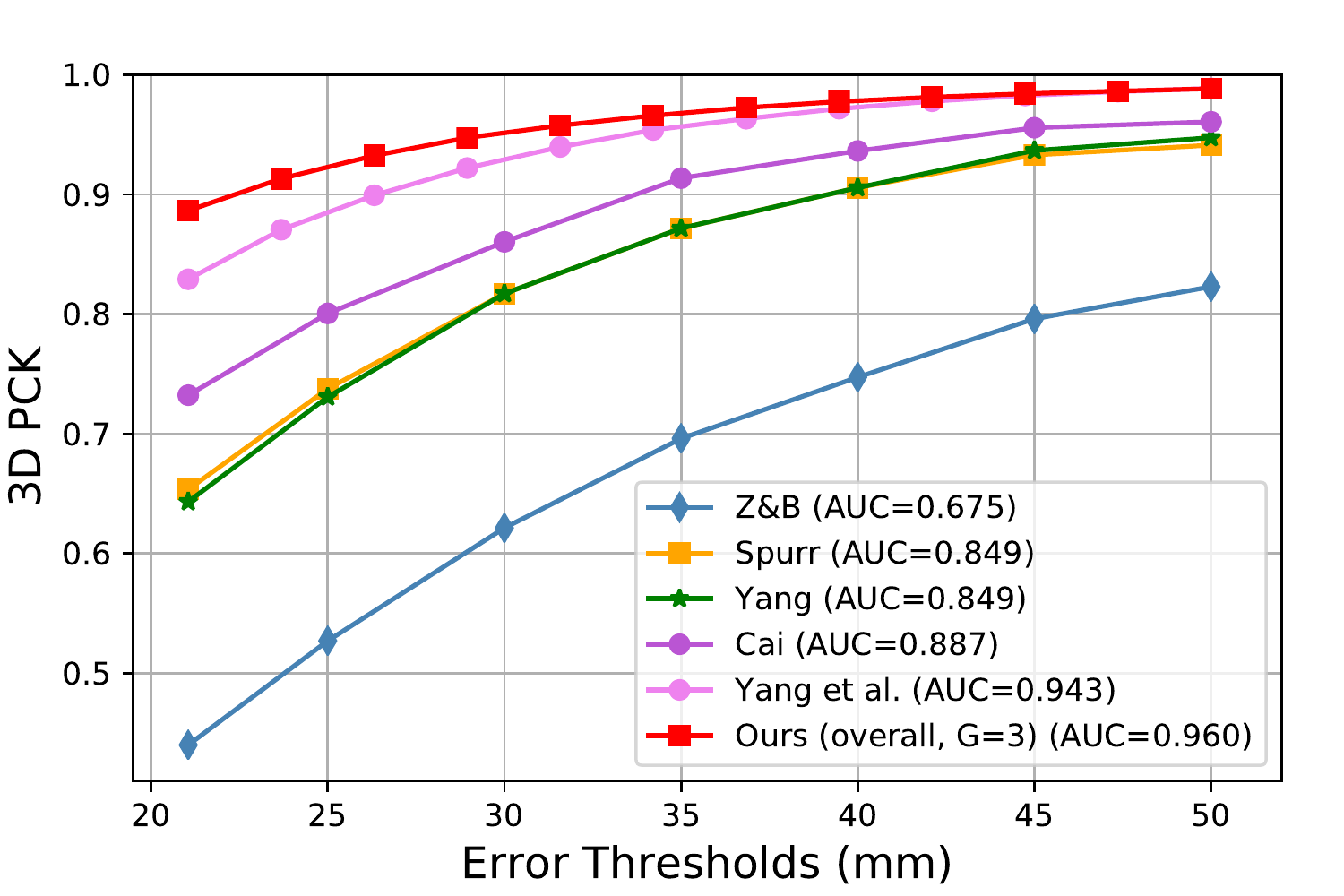}
}
\subfigure[STB]{
\label{Fig.AUC_Sota.2}
\includegraphics[width=0.3\textwidth]{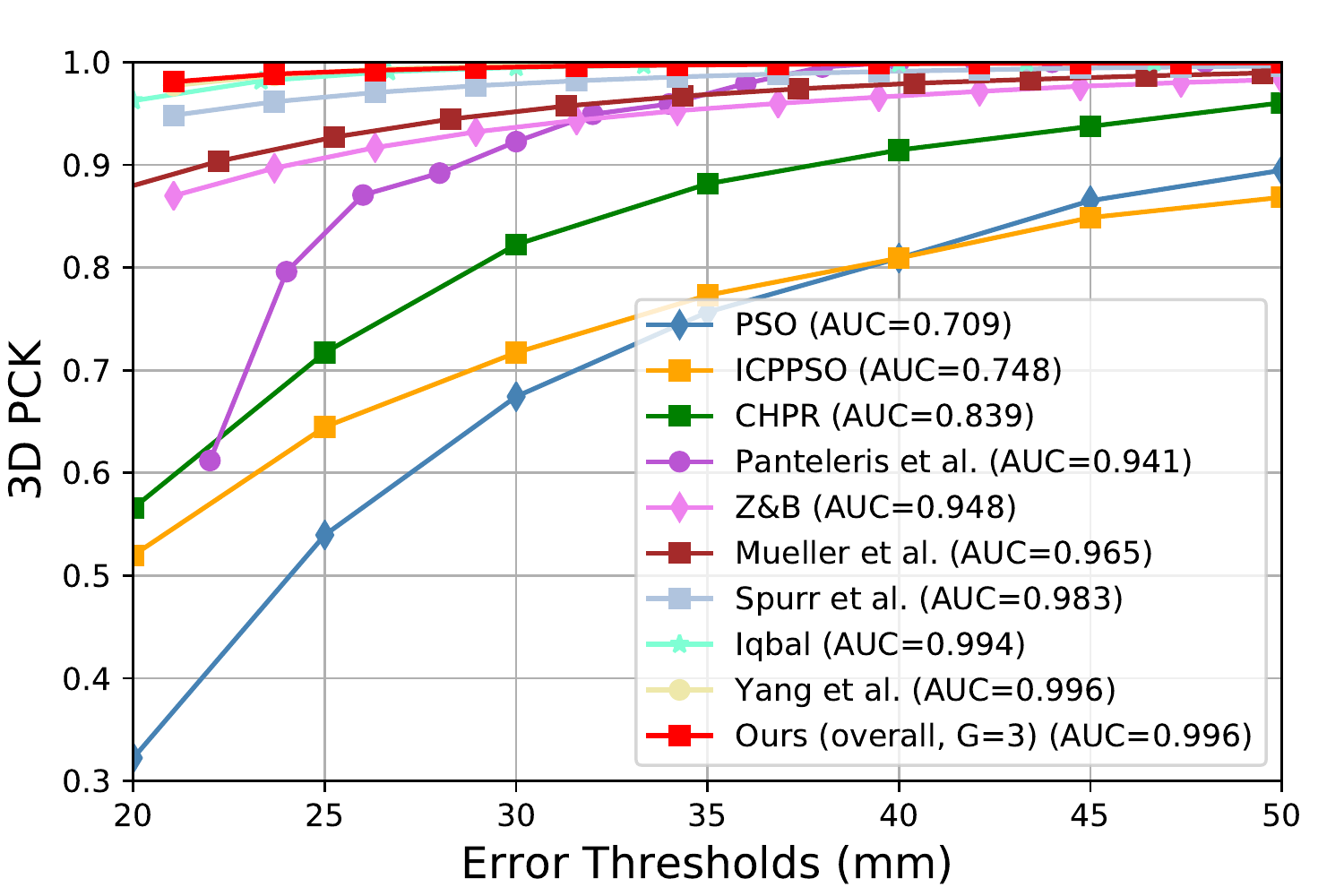}
}
\subfigure[Dexter + Object]{
\label{Fig.AUC_Sota.3}
\includegraphics[width=0.3\textwidth]{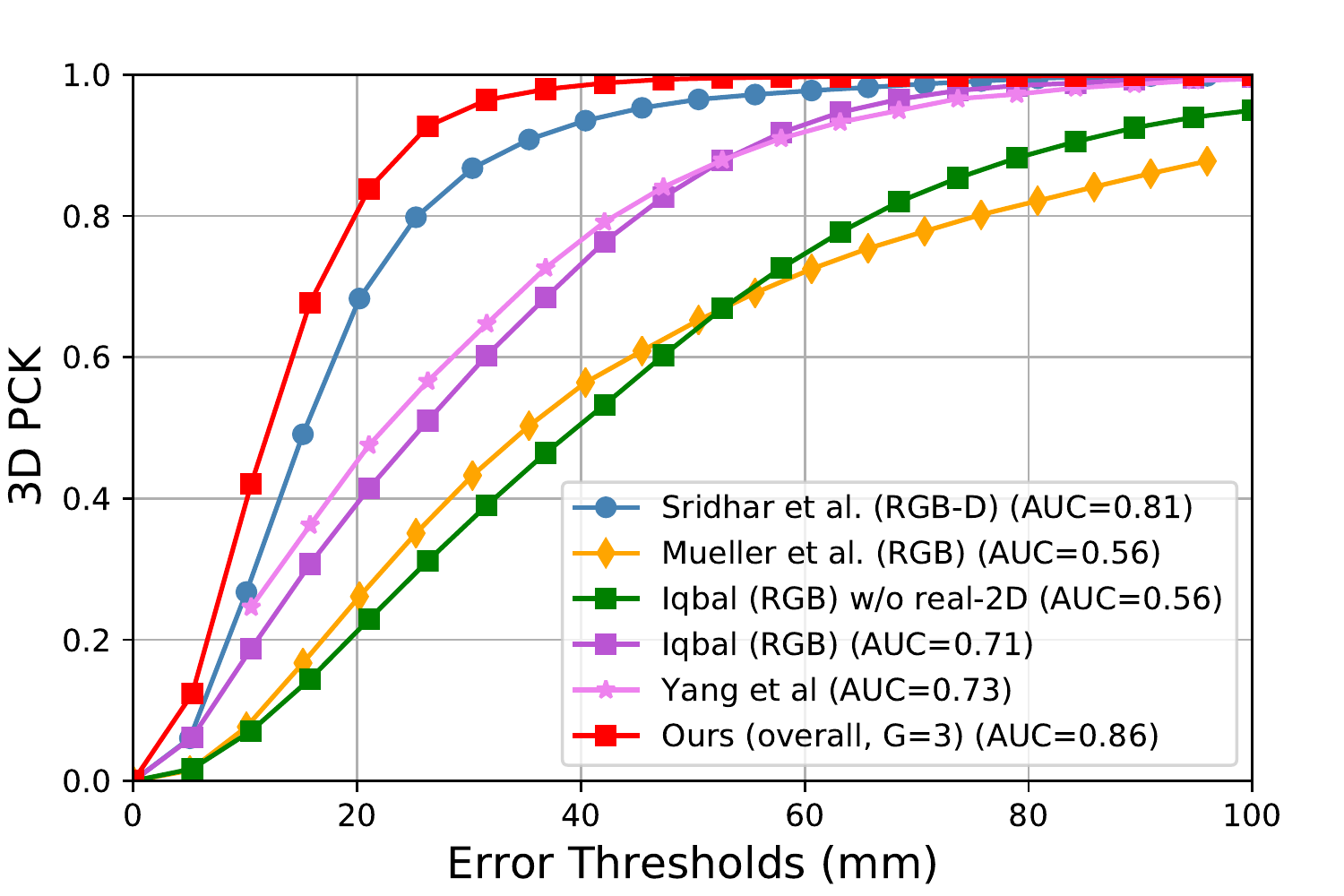}
}
\vspace{-2mm}
\caption{Comparison with SOTA methods on the RHD, the STB, and the Dexter + Object datasets. For the RHD dataset, the results of some methods (\eg, Kulon et al. \cite{dkulon2020cvpr}, Iqbal et al. \cite{iqbalLatent25D}) are not shown here because their AUC curves of this dataset are not available yet. (Best view in colors.)}
\label{Fig.AUC_Sota}
\end{figure*}
\subsection{Implementation Details}

\noindent {\bf Data Processing.} We implement the data pre-processing and augmentation similar to Yang et al. \cite{AligningHandICCV2019}. Specifically, we first crop original RGB images using the bounding box calculated by the ground truth masks and resize the cropped image to $256\times 256$. Then, we apply an online data augmentation with a random scaling between $[1, 1.2]$, a random rotation between $[-\pi, \pi]$, a random translation between $[-20, 20]$, and a color jittering with a random hue between $[-0.1, 0.1]$.
As for the global hand scale and root joints, we follow the same way as Yang et al. \cite{AligningHandICCV2019} using the MCP of the middle finger as the root joint and selecting the euclidean distance between the MCP and the PIP of the middle finger as the global scale for fair comparisons. We also align the annotations of the RHD and the STB datasets so that the joints with the same annotation index have the same semantic meaning.

\noindent {\bf Optimization.} We use Adam optimizer \cite{kingma2014adam} to train the network. We train the \emph{shared feature extraction module} (hourglass network) to predict 2D joints location using $\ell_1$ loss for initialization, with a learning rate of 1e-3 and a mini-batch size of 64. Then, we use the loss function defined in Eq. \eqref{loss1} with $\beta=20$ to optimize the overall network. The learning rates for the newly introduced \emph{grouped feature learning module} and \emph{feature fusing module} are 1e-1 and 1e-2, respectively. For the remaining network parameters, the learning rate is set to 1e-4 with a mini-batch size of 32.

For training, we initialize every $\theta_{i, j}$ of Eq. \eqref{concrete} to be $1/K$ as we do not impose priors for the group categorization ($K$ is the number of groups). $\tau$ of Eq. \eqref{concrete} is initialized to be 5 and decrease 0.1 for every 1,000 steps until it reaches around 0. We use the number of groups as 3 (i.e., $K = 3$) in all of our experiments as we find that further increasing the number of groups produces comparable results (as shown in our ablation analysis in the main paper), which coincides with the conclusion from \cite{YingWuCVPR2019}.

\subsection{Benchmark Results}

We compare our proposed method with the SOTA methods \cite{sridhar2016real,z&B2017,cai2018weakly,mueller2018ganerated,panteleris2018using,spurr2018cross,iqbalLatent25D,yang2019disentangling,AligningHandICCV2019}, on various benchmark datasets to illustrate the effectiveness of our proposed method. The performances of the SOTA methods on each dataset are obtained from their original paper. 

As shown in Fig. \ref{Fig.AUC_Sota.1}, our method outperforms the SOTA methods \cite{z&B2017,spurr2018cross,yang2019disentangling,cai2018weakly,AligningHandICCV2019} by a large margin on the RHD dataset, which demonstrates the promising performance of our proposed method. Comparing with \cite{iqbalLatent25D} (AUC [20-50] is 0.94), and \cite{dkulon2020cvpr} (AUC [20-50] is 0.956), our proposed method (AUC [20-50] is 0.96) also has better performance. 

Our method gets comparable results on the STB dataset with the SOTA methods 
\cite{panteleris2018using,z&B2017,mueller2018ganerated,spurr2018cross,iqbalLatent25D,AligningHandICCV2019,cai2019exploiting} without using additional training data. Note that several works such that \cite{cai2019exploiting, Baek_2020_CVPR, Theodoridis_2020_CVPR_Workshops} (AUC [20-50] are 0.995, 0.995, 0.997, respectively) are not included in Fig. \ref{Fig.AUC_Sota.2} because their AUC curves are not available yet. Apart from less training data used, such comparable results are also partly because the STB dataset is a much easier dataset, where many algorithms perform considerably well. 

For the Dexter + Object dataset, we follow the same training setup as that used in \cite{z&B2017,mueller2018ganerated,yang2019disentangling,AligningHandICCV2019}, which makes use of both the RHD and the STB datasets for training. But note that our method does not include extra augmentation of objects. Besides, our evaluation on this dataset is the same as \cite{AligningHandICCV2019} which adopts the best root and scale. Following previous methods, the AUC [0-50] is reported, and the comparison is shown in Fig. \ref{Fig.AUC_Sota.3}.

\begin{table*}[ht]
\centering
\fontsize{9pt}{0.9\baselineskip}\selectfont
\begin{tabular}{@{\extracolsep{\fill}} l | c | c c  }
\hline
\hline 
Methods & Use Shape & Mean EPE [mm] ($\downarrow$) & AUC [0-100] ($\uparrow$) \\
\hline
Boukh. et al. \cite{CVPR2019_Mesh_inthewild} & \checkmark & 3.5 & 0.351 \\
Hasson et al. \cite{hasson2019learning} &  \checkmark & 1.33 & 0.737 \\
MANO Fit \cite{FreiHAND2019}& \checkmark & 1.37 & 0.73 \\
MANO CNN \cite{FreiHAND2019}&  \checkmark & 1.1 & 0.783 \\
Kulon et al. \cite{dkulon2020cvpr} & \checkmark & 0.84 & 0.834 \\
\hline 
Baseline (\textbf{ours}) & -   &  0.932 & 0.816  \\
+ Group Learned. & - & 0.862 & 0.830  \\ 
+ Fea. Fuse.  & - & 0.856 & 0.831 \\
+ \textbf{Ensemble} & - & \textbf{0.796} & \textbf{0.842} \\
\hline 
\hline
\end{tabular}
\vspace{-1mm}
\caption{Results (aligned \cite{schonemann1966generalized}) on the FreiHAND dataset, where baseline means not to learn joint groups (\ie, the number of groups is 1). Fea. Fuse means whether to learn the (unique) feature embedding for each group exploiting the features from all the groups. Our checkpoints ensemble \cite{Liu_2018_ECCV_Workshops} result reaches the best performance. $\uparrow$/$\downarrow$ represents the higher/lower the better.}
\label{Table:Ablation_FreiHAND}
\end{table*}
To further demonstrate the effectiveness of our method, we also show the detailed results on the FreiHAND dataset in Table \ref{Table:Ablation_FreiHAND}. 
For the FreiHAND competition, our method ranks Top-1 over the submissions without exploiting the dense 3D shape label at the submission date (see Supp.).

\subsection{Ablation Analysis}

\begin{table*}[ht]
\centering
\fontsize{9pt}{0.9\baselineskip}\selectfont
\begin{tabular}{@{\extracolsep{\fill}} c c c | c  c |c }
\hline
\hline
\multicolumn{2}{c}{Group} & \multirow{2}{*}{Fea. Fuse} & \multirow{2}{*}{Mean EPE [mm] ($\downarrow$)} & \multirow{2}{*}{Median EPE [mm] ($\downarrow$)} & \multirow{2}{*}{AUC [20-50] ($\uparrow$)} \\
Manual & Learned & & & & \\
\hline
- & -  & - & 11.534  & 8.885  & 0.951  \\
\checkmark & - & - & 11.205  & 8.622  & 0.954 \\
\checkmark & - & \checkmark & 11.021 & 8.543  & 0.957 \\ 
- & \checkmark & - & 10.771 & 8.283 & 0.958 \\
- & \checkmark & \checkmark & \textbf{10.653} & \textbf{8.200} & \textbf{0.960}  \\

\hline
\hline
\end{tabular}
\vspace{-1mm}
\caption{Ablation analysis of the proposed method on the RHD dataset, where Fea. Fuse means whether to learn the feature embedding for each group exploiting the features from all groups.}
\label{Table:AblationRHD}
\end{table*}

We perform detailed ablation analysis mainly on the RHD dataset to investigate each proposed component of the grouped feature learning module. 

Firstly, we are especially interested in investigating that i) \emph{whether categorizing the hand joints into groups improves the performance}? If so, ii) \emph{how about grouping the joints manually}? Moreover, iii) \emph{whether further learning a unique feature embedding for each group, by exploiting the features from all the groups, is effective}? We validate those hypotheses in Table \ref{Table:AblationRHD}, where the manual grouping of joints is similar to the idea of \cite{YingWuCVPR2019} and we separate the hand joints into three groups (\ie, thumb, index, and other fingers) following Zhou et al. \cite{zhou2018hbe}. Table \ref{Table:AblationRHD} illustrates that the proposed method, with learnable joints grouping and feature fusing among all the groups, achieves the best performance. The improvements shown in Table \ref{Table:Ablation_FreiHAND} further illustrate the effectiveness of the proposed modules.

After validating the effectiveness of the proposed method, the next immediate question is \emph{how many groups should we categorize the joints into}? In Table \ref{Table:GroupNumber}, we show the performance when learning 2-5 groups, which demonstrates that learning larger than 3 groups produces comparable results, coinciding with the conclusion from \cite{YingWuCVPR2019}. 

\begin{figure}[ht]
\centering
\subfigure[RHD]{
\label{Fig.GroupRes.1}
\includegraphics[width=0.13\textwidth]{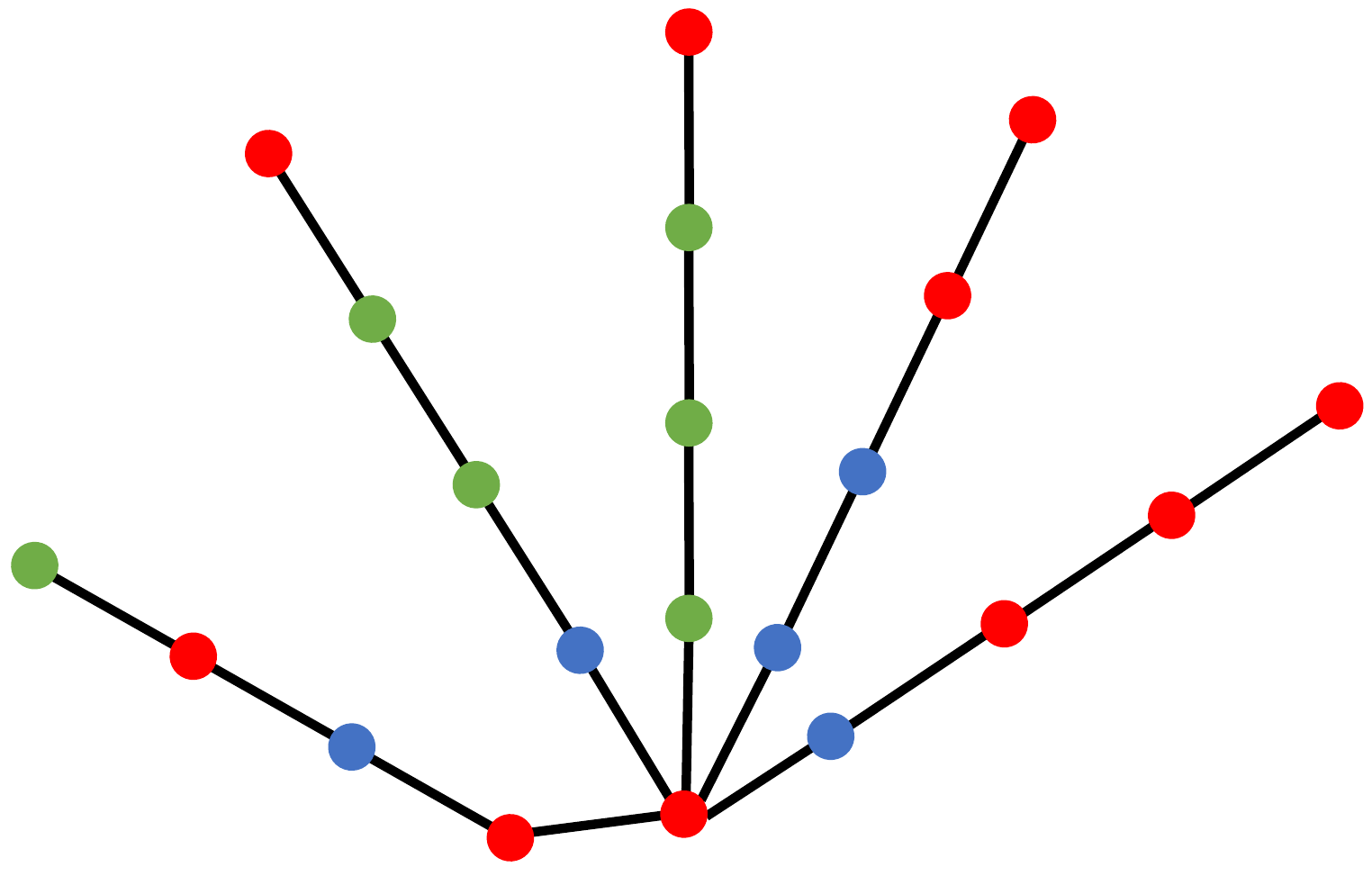}} \hspace{2mm}
\subfigure[STB]{
\label{Fig.GroupRes.2}
\includegraphics[width=0.13\textwidth]{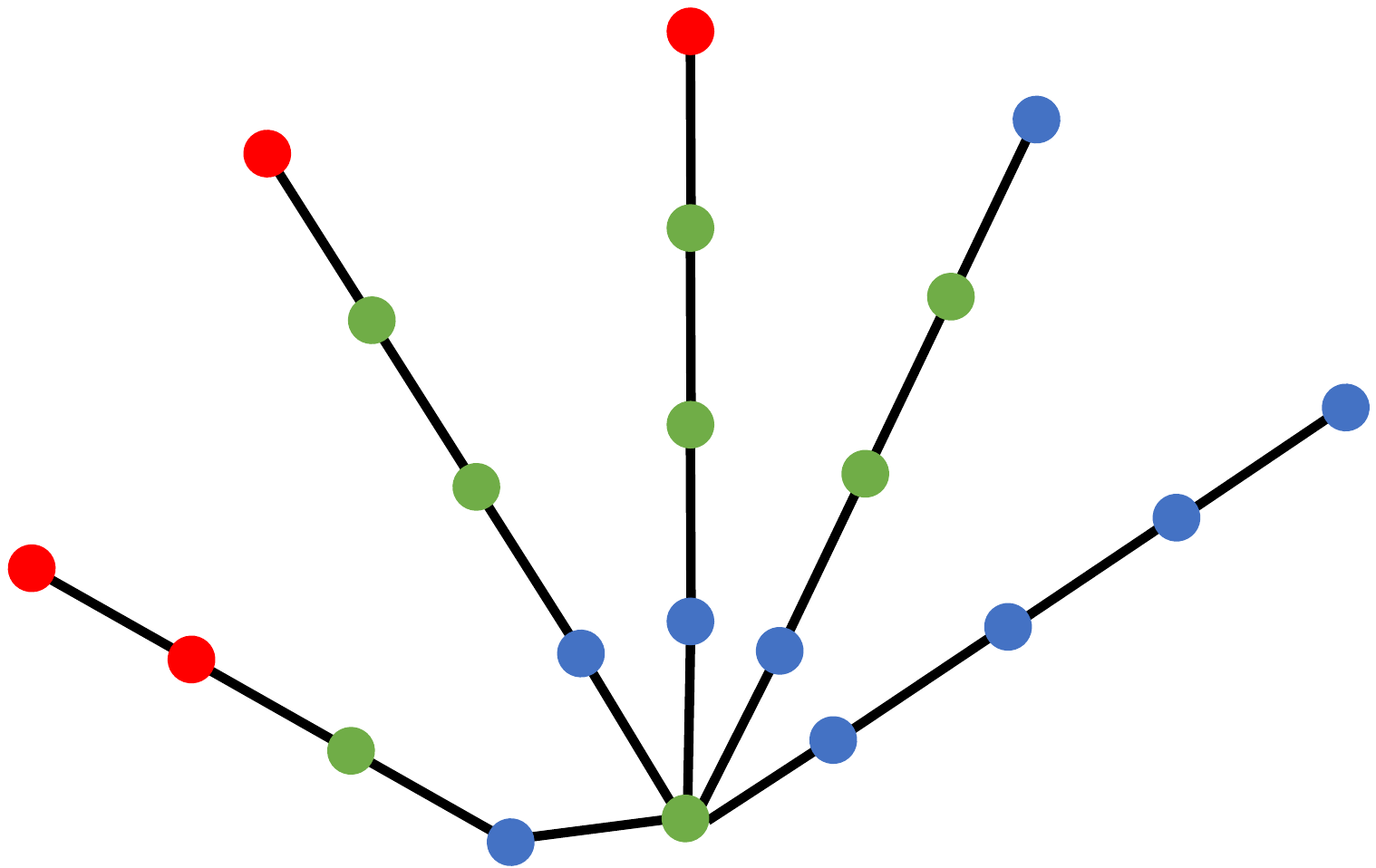}} \hspace{2mm}
\subfigure[FreiHAND]{
\label{Fig.GroupRes.3}
\includegraphics[width=0.13\textwidth]{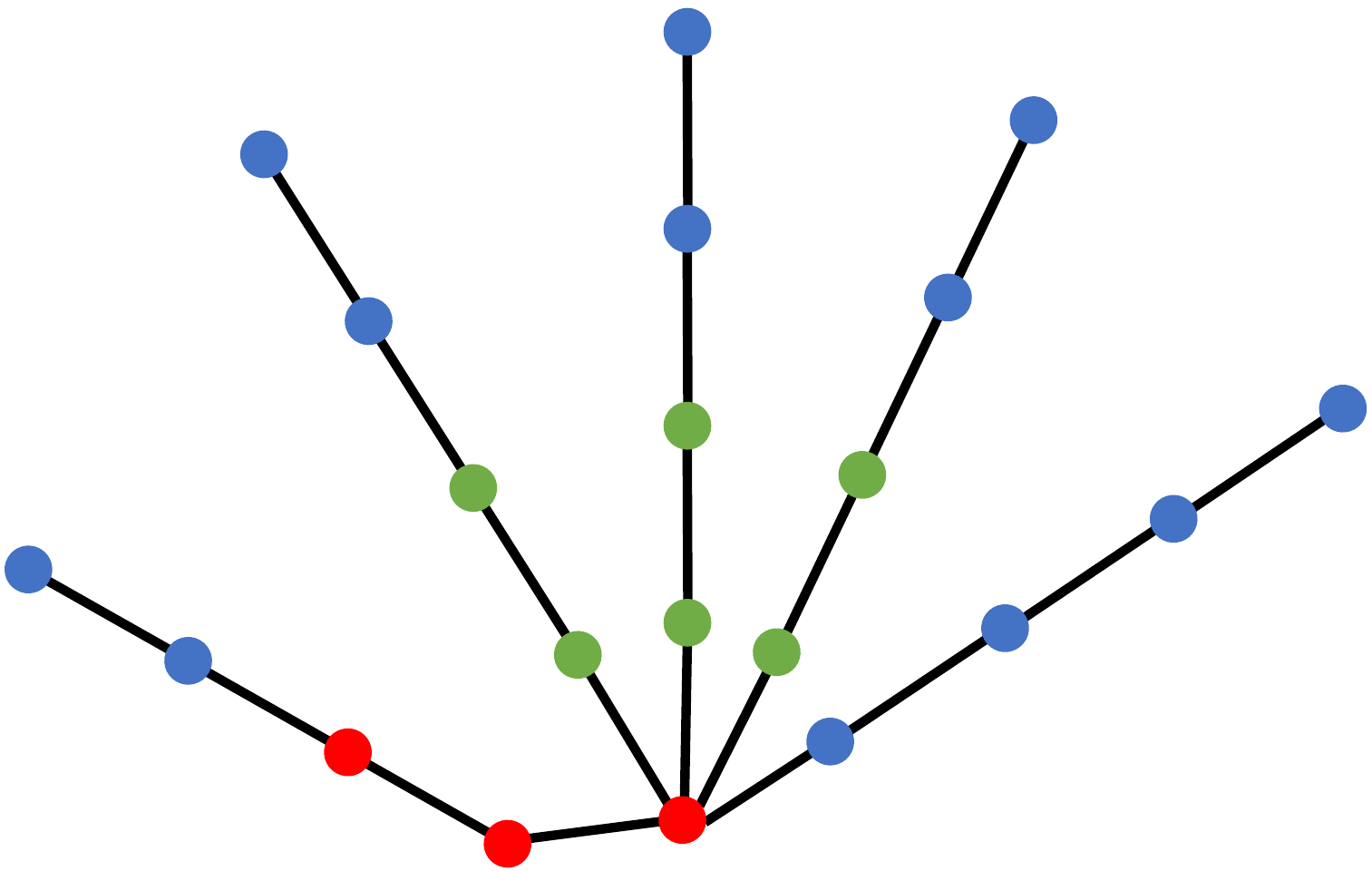}}
\vspace{-2mm}
\caption{The automatically learned joints grouping results on the RHD, the STB, and FreiHAND datasets (with the number of groups equal to 3). The red, green, and blue represent the three groups, respectively. (Best view in colors.)}
\label{Fig.GroupRes}
\end{figure}

The grouping results (shown in Fig.\ref{Fig.GroupRes}) validate that the joints cluster is not only related to constraints due to the skeletons and nerves, but also related to the distribution of poses in the datasets \cite{jahangiri2017generating,YingWuCVPR2019}. For example, one dataset may contain more hand images from a kitchen scenario, while another is collected from sporting scenarios, intuitively they should have distinctive grouping results, and our method can well characterize those differences. We have verified the good transferring-ability of the learned groups to a different dataset, please see Fig. \ref{Fig.AUC_Sota.3}, where the model is trained on the RHD and the STB datasets and evaluated on the Dexter + Object dataset.

\begin{table}[h]
\centering
\fontsize{9pt}{0.9\baselineskip}\selectfont
\begin{tabular}{@{\extracolsep{\fill}} c| c c | c  c |c }
\hline
\hline
$\#\text{Group}$ & Mean EPE ($\downarrow$) & Median EPE ($\downarrow$) & AUC [20-50] ($\uparrow$) \\
\hline
Baseline & 11.534  & 8.885  & 0.951  \\
\hline
2 & 11.042  & 8.478  & 0.956  \\ 
\textbf{3} & \textbf{10.771} & \textbf{8.283} & \textbf{0.958} \\
4 & 10.808 & 8.340 & 0.958  \\ 
5 & 10.923 & 8.418 & 0.957  \\ 
21 & 10.964 & 8.422 & 0.956 \\ 
\hline
\hline
\end{tabular}
\vspace{-1mm}
\caption{Ablation analysis of the number of groups ($\#\text{Group}$) on the RHD dataset, where baseline means not to learn joint groups (\ie, $\#\text{Group}= 1$). The Mean/Median EPE are in mm.}
\label{Table:GroupNumber}
\end{table}

\section{Conclusions \label{conclusion}}
We propose a novel method for 3D hand pose estimation from a single RGB image, which automatically categorizes the hand joints into groups. By treating groups as different tasks that learn different features, our method efficiently avoids negative transfer across groups. Moreover, we further exploit features from different groups to learn a more discriminative feature embedding for each group. We carried out extensive experiments and detailed ablation analysis to illustrate the effectiveness of our method and the proposed network can be optimized end-to-end in deep neural networks. Our results on the RHD, the STB, the Dexter + Object, and the FreiHAND datasets significantly outperform the SOTA methods.

\section*{Broader Impact}
Our work improves hand pose estimation performance which can facilitate much easier animation fabrication, more accurate sign language recognition, and many other human-computer interaction applications. All of those contribute a lot to us especially carton filmmakers and physically challenged people. What's more, our work has some positive impacts on the academic research community. Our proposed method can be easily applied to many other tasks such as human pose estimation, hand objects estimation, and other related tasks. 

Since our method does not use the identity information of each individual, the authors believe there is no offensive to ethical or personal privacy.

\section*{Acknowledgments}

This work was supported by the National Natural Science Foundation of China under grant 61871435 and the Fundamental Research Funds for the Central Universities no. 2019kfyXKJC024.

\bibliography{Grouping_Final}
\end{document}